\documentclass[letterpaper, 10 pt, conference]{ieeeconf}  % Comment this line out
\IEEEoverridecommandlockouts                              % This command is only
\usepackage{amsmath}     % for math environments
\usepackage{amssymb}     % for math symbols
\usepackage{graphicx}    % for including images    % for hyperlinks
\usepackage{booktabs}    % for nicer tables
\usepackage{xcolor}  
\usepackage{amsmath, amssymb}
\usepackage{graphicx}
\usepackage{tikz}
\usepackage{booktabs}
\usepackage{xcolor}
\usepackage{bbm}
\usepackage{algorithm}
\usepackage{algorithmic}
\usepackage{booktabs}
\usepackage{xcolor}
\usepackage{url} 
\usepackage[table]{xcolor}
\usepackage[table]{xcolor}
\definecolor{localgrey}{RGB}{150,150,150}
\definecolor{fedavgblue}{RGB}{0,112,192}
\definecolor{globaldark}{RGB}{30,30,30}

\title{\LARGE \bf
Federated Learning for Financial Forecasting
}

\author{Manuel Noseda\\
ETH Zürich\\
{\tt\small mnoseda@ethz.ch}
\and
Alberto De Luca\\
ETH Zürich\\
{\tt\small adeluc@ethz.ch}
\and
Lukas Von Briel \\
ETH Zürich\\
{\tt\small lvonbriel@ethz.ch}
\and
Nathan Lacour \\
ETH Zürich\\
{\tt\small nlacour@ethz.ch}
}

\begin{document}

\maketitle
\thispagestyle{empty}
\pagestyle{empty}

%%%%%%%%%%%%%%%%%%%%%%%%%%%%%%%%%%%%%%%%%%%%%%%%%%%%%%%%%%%%%%%%%%%%%%%%%%%%%%%%
\begin{abstract}

This paper studies Federated Learning (FL) for binary classification of volatile financial market trends. Using a shared Long Short-Term Memory (LSTM) classifier, we compare three scenarios: (i) a centralized model trained on the union of all data, (ii) a single-agent model trained on an individual data subset, and (iii) a privacy-preserving FL collaboration in which agents exchange only model updates, never raw data. We then extend the study with additional market features, deliberately introducing not independent and identically distributed data (non-IID) across agents, personalized FL and employing differential privacy. Our numerical experiments show that FL achieves accuracy and generalization on par with the centralized baseline, while significantly outperforming the single-agent model.
The results  show that collaborative, privacy-preserving learning provides collective tangible value in finance, even under realistic data heterogeneity and personalization requirements.

\vspace{1em}
Index Terms---financial forecasting, federated learning, deep learning, privacy, consensus, volatility.

\end{abstract}

%%%%%%%%%%%%%%%%%%%%%%%%%%%%%%%%%%%%%%%%%%%%%%%%%%%%%%%%%%%%%%%%%%%%%%%%%%%%%%%%
\section{Introduction}

Accurately forecasting market prices is still one of the most fundamental and challenging tasks in finance.  It represents the backbone of portfolio optimization and studies high-frequency time-series data that are notoriously volatile. Indeed, return distributions are heavy-tailed and heteroskedastic, exhibiting volatility clustering. Large (or small) returns in magnitude are likely to be followed by similarly large (or small) returns, though not necessarily in the same direction. This allows short-term risk forecasting but makes directional trend prediction inherently difficult \cite{cont2001empirical}.
However, in real-world settings, data pooling can be very difficult, since it is often distributed among multiple entities and privacy or regulatory restrictions can prevent data centralization. Federated Learning (FL), a novel distributed machine learning technique presented in \cite{mcmahan2017communication}, offers a framework for collaborative model training without the need for raw data sharing, thus preserving privacy.

While FL is well established in sensitive domains such as healthcare \cite{rieke2020future}, smart energy microgrids \cite{preethi2021flsmartgrid} and large-scale recommender systems \cite{zhang2022fedrec},  its benefits and limitations for financial forecasting and classification remain relatively under-explored \cite{wef2020federatedfinance}. 
Imagine a network of n agents, such as, banks, asset-management firms, or exchanges, each holding a proprietary, partial view of the same market . The agents wish to jointly train a model that predicts future asset returns and steers portfolio-allocation decisions in order to pursue  higher utility for everyone. However, non-disclosure agreements, client confidentiality and competitive pressures make raw data pooling infeasible  \cite{eba2019outsourcing}, \cite{gdpr2016}.

In the following, we answer a single, practical question: if direct data sharing is off the table, can agents still exchange only the minimal information necessary, securely and efficiently, to capture the performance improvements that come from collaboration?  We benchmark federated learning (FL) against single‑agent and centralized baselines, quantify the privacy–communication trade‑off and evaluate robustness under data that are not independent and identically distributed (non‑IID). 
 
The remainder of the paper is organized as follows: Section II reviews the mathematical preliminaries; Section III formalizes the learning problem; Section IV reports the numerical experiments and simulations; Section V closes with the main findings and directions for future research.

\section{Mathematical Preliminaries}
In this section we review fundamental definitions of graph theory and optimization that will come in handy when we will make assumptions for our model to guarantee convergence of the consensus network structure. We refer to \cite{bullo2023lectures} and \cite{boyd2004convex}
for a deeper discussion of the fundamentals.
\subsection{Graph Theory}
Let $\mathcal G = (\mathcal V,\mathcal E)$ be a directed communication graph with node set $\mathcal V \;=\;\{\,1,2,\dots,N\,\}$, where an ordered edge $(i,j)\in\mathcal E$ indicates that client $i$ can transmit information to client $j$. The adjacency matrix $A=[A_{ij}]\in\mathbb{R}^{N\times N}$ is defined by $A_{ij}>0$ only if $(i,j)\in\mathcal{E}$ or $i=j$  (and $A_{ij}=0$ otherwise). 
$A$ is  row-stochastic when $\sum_{j=1}^{N} A_{ij}=1\;\forall i$ and doubly-stochastic  when, in addition, $\sum_{i=1}^{N} A_{ij}=1\;\forall j$.
A directed graph $\mathcal G$ is \emph{strongly connected} when, for every
ordered pair of nodes $(i,j)$, there exists a directed path from $i$ to $j$.
Strong connectivity ensures that information injected at any client can
eventually reach all others through the mixing process.

\subsection{Optimization}
A differentiable function $g:\mathbb{R}^{p}\!\to\!\mathbb{R}$ is \emph{convex} if $g(\lambda u+(1-\lambda)v)\le\lambda g(u)+(1-\lambda)g(v)$ for all $u,v \in \mathbb{R}^p$ and $\lambda\in[0,1]$.
It is \emph{$L$-Lipschitz smooth} when$\|\nabla g(u)-\nabla g(v)\|\le L\|u-v\|$  for all $u,v \in \mathbb{R}^p$. Each client $i$ owns a differentiable convex local loss $f_i:\mathbb{R}^p\to\mathbb{R}$ and the global separable objective function is $F(\mathbf{x}) = \tfrac{1}{N} \sum_{i=1}^N f_i(\mathbf{x})$ for $\mathbf{x} \in \mathbb{R}^p$.
For client $i$ we assume access to an unbiased stochastic gradient (SG) $\hat{\mathbf{g}}_i(\mathbf{x})$ satisfying
$\mathbb{E}[\hat{\mathbf{g}}_i(\mathbf{x})] = \nabla f_i(\mathbf{x}), \quad
\mathbb{E}\left[\lVert \hat{\mathbf{g}}_i(\mathbf{x}) - \nabla f_i(\mathbf{x}) \rVert^{2}\right] \le \sigma^{2}$, for some finite variance bound $\sigma^2$.

\section{Problem Formulation}
\label{sec:problem-formulation}

Financial institutions hold rich, yet fragmented streams of market data, but they share the common objective of predicting market behaviour.  The following problem formulation aims to find a binary output description of the market trend.
We consider a Federated Learning (FL) network of \(N\) clients, each with a private local dataset
\[
  \mathcal{D}_i = \{(x_{i,j}, y_{i,j})\}_{j=1}^{n_i},
  \qquad
  K = \sum_{i=1}^{N} n_i,
\]
where \(x_{i,j} \in \mathbb{R}^d\) are feature vectors (e.g., cumulative log-returns and temporal lags)  and \(y_{i,j} \in \{0,1\}\) (e.g., 1 if the market is going up, 0 otherwise) are binary trend labels.  
The collaborative goal is to find model parameters \(w \in \mathbb{R}^d\) that minimize the global empirical risk
\[
  F(w) \;=\; \frac{1}{K}\sum_{i=1}^{N}\sum_{j=1}^{n_i}
  \ell\!\bigl(f(w;\,x_{i,j}),\,y_{i,j}\bigr),
\]
with model output $f(w;\,x_{i,j}) \in [0,1]$.

For binary classification, we adopt the binary cross-entropy loss:
\[
  \ell(\tilde{y}, y) = - y \log \tilde{y} - (1 - y) \log(1 - \tilde{y})
\]
\[
  \mathrm{Acc}(w) = \frac{1}{|\mathcal{D}_{\text{test}}|}
  \sum_{(x,y)\in\mathcal{D}_{\text{test}}}
  \mathbf{I}\{ y = \hat{y} \},
\]
\[
  \text{where } \hat{y} = \mathbf{I}\bigl[ f(w;\,x) \ge 0.5 \bigr],
\]

where  \( \mathbf{I}(\cdot) \) is the indicator function and thresholds the predicted probability at \(0.5\).
All clients agree on a common model architecture for forecasting, and we choose a small-size Long Short-Term Memory (LSTM), widely used in binary financial forecasting literature \cite{fjellstrom2022lstm}.

\medskip
Following the procedure in \cite{mcmahan2017communication}, we deploy the Federated Averaging (FedAvg) algorithm:
\begin{algorithm}
\caption{FedAvg}
\label{alg:fedavg}
\begin{algorithmic}[1]
\STATE \textbf{Input:} Number of rounds $T$, number of clients $K$, local epochs $E$
\STATE Initialize global model weights $w^{(0)}$
\FOR{each round $t = 1$ to $T$}
    \FOR{each client $k \in K$ \textbf{in parallel}}
        \STATE Initialize $w_k^{(t)} \leftarrow w^{(t-1)}$
        \FOR{each local epoch $e = 1$ to $E$}
            \STATE Update $w_k^{(t)}$ using SG descent algo on local data
        \ENDFOR
        \STATE Send updated weights $w_k^{(t)}$ to server
    \ENDFOR
    \STATE \textbf{Server aggregates:} 
    $w^{(t)} \leftarrow \frac{1}{K} \sum_{k \in K} w_k^{(t)}$
\ENDFOR
\STATE \textbf{Output:} Final global model weights $w^{(T)}$
\end{algorithmic}
\end{algorithm}

\subsection{Consensus and Network Structure}

\begin{figure}[H]
    \centering
\begin{tikzpicture}[>=latex, node distance=2cm]

%---- Step 1 label ----%
\node[align=center] (step1) at (0,2.1) {\textbf{Step 1}};

%---- Left Figure ----%
\node[draw, minimum size=0.9cm, shape=rectangle, thick] (squareL) at (0,0) {Server};
\node[draw, circle, minimum size=0.7cm, thick] (A) at (90:1.4) {1};
\node[draw, circle, minimum size=0.7cm, thick] (B) at (210:1.4) {2};
\node[draw, circle, minimum size=0.7cm, thick] (C) at (330:1.4) {3};

\draw[->, thick] (A) -- node[pos=0.55, right] {\scriptsize $\Delta w_1$} (squareL);
\draw[->, thick] (B) -- node[pos=0.55, above left] {\scriptsize $\Delta w_2$} (squareL);
\draw[->, thick] (C) -- node[pos=0.55, above right] {\scriptsize $\Delta  w_3$} (squareL);

%---- Step 2 and Right Figure ----%
\begin{scope}[xshift=4cm] % Increased xshift for more spacing

\node[align=center] (step2) at (0,2.1) {\textbf{Step 2}};

\node[draw, minimum size=0.9cm, shape=rectangle, thick] (squareR) at (0,0) {Server};
\node[draw, circle, minimum size=0.7cm, thick] (A2) at (90:1.4) {1};
\node[draw, circle, minimum size=0.7cm, thick] (B2) at (210:1.4) {2};
\node[draw, circle, minimum size=0.7cm, thick] (C2) at (330:1.4) {3};

\draw[->, thick] (squareR) -- node[pos=0.55, right] {\scriptsize $w$} (A2);
\draw[->, thick] (squareR) -- node[pos=0.55, above left] {\scriptsize $w$} (B2);
\draw[->, thick] (squareR) -- node[pos=0.55, above right] {\scriptsize $w$} (C2);

\end{scope}

\end{tikzpicture}
\caption{Overview of FedAvg process with 3 agents ($\Delta w_i$ represents the locally updated model of the i-th agent)}
\label{fig:serv-steps}
\end{figure}
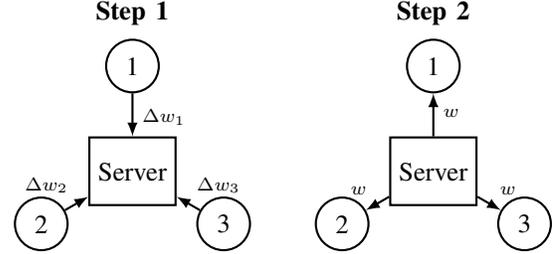

In the real network, each agent is directly linked to the global server and is not connected to any other agent in the network. The global server is the main tool that is going to enable FedAvg to work. It receives model updates from all agents before aggregating them and sending back the same updated model to all agents (see Fig. \ref{fig:serv-steps}).

However, the communication structure in FL can be abstracted as a weighted directed graph $G = (V, E)$, where nodes represent agents, and edges represent communication links. 
While this network structure appears to represent direct peer-to-peer communication among all agents, it is not the case in reality. Indeed, as mentioned before, in our implementation all aggregation is performed by a static central server (each agent sends its model update to the server, which computes the average and redistributes the result). The central server is thus not a node of the graph, but is implicitly represented by the edges (i.e,  the aggregation step corresponds mathematically to synchronous averaging over this complete graph).

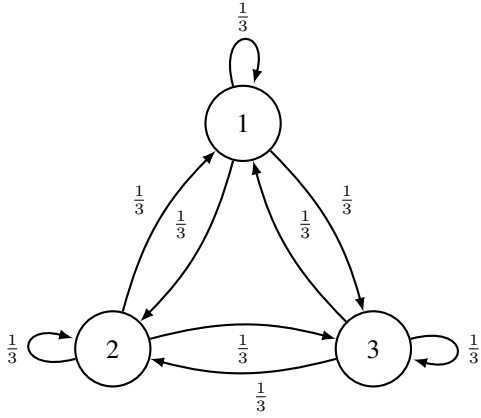
\begin{figure}[htb]
    \centering
    \begin{tikzpicture}[>=latex, node distance=2cm]
        % Nodes
        \node[draw, circle, minimum size=1cm, thick] (A) at (90:2) {1};
        \node[draw, circle, minimum size=1cm, thick] (B) at (210:2) {2};
        \node[draw, circle, minimum size=1cm, thick] (C) at (330:2) {3};
        
        % Edges between nodes
        \draw[->, thick] (A) to [bend left=15] node[above left,pos=0.5] {\small $\frac{1}{3}$} (B);
        \draw[->, thick] (B) to [bend left=15] node[below,pos=0.5] {\small $\frac{1}{3}$} (C);
        \draw[->, thick] (C) to [bend left=15] node[above right,pos=0.5] {\small $\frac{1}{3}$} (A);
        
        % Reverse edges for bidirectionality (optional, comment out if not needed)
        \draw[->, thick] (B) to [bend left=15] node[above  left,pos=0.5] {\small $\frac{1}{3}$} (A);
        \draw[->, thick] (C) to [bend left=15] node[below right,pos=0.5] {\small $\frac{1}{3}$} (B);
        \draw[->, thick] (A) to [bend left=15] node[above right,pos=0.5] {\small $\frac{1}{3}$} (C);
        
        % Self-loops
        \draw[->, thick] (A) edge [loop above] node[above] {\small $\frac{1}{3}$} ();
        \draw[->, thick] (B) edge [loop left] node[left] {\small $\frac{1}{3}$} ();
        \draw[->, thick] (C) edge [loop right] node[right] {\small $\frac{1}{3}$} ();
    \end{tikzpicture}
    \caption{Weighted directed graph for 3 agents}
    \label{fig:three-circles}
\end{figure}

In our approach, the $N$ agents are modeled as nodes in a strongly connected directed graph, where each agent (node) is connected to every other agent, including itself, via an edge of equal weight. Specifically, the adjacency (connectivity) matrix $A \in \mathbb{R}^{N\times N}$ is defined by
\[
A_{ij} = \frac{1}{N}, \qquad \forall i, j = 1,\ldots,N
\]
For example, in the case of $N=3$ agents (see Fig. \ref{fig:three-circles}), the adjacency matrix $A$ is
\[
A = \begin{bmatrix}
\frac{1}{3} & \frac{1}{3} & \frac{1}{3} \\
\frac{1}{3} & \frac{1}{3} & \frac{1}{3} \\
\frac{1}{3} & \frac{1}{3} & \frac{1}{3}
\end{bmatrix}
\] 
This means that $A$ is symmetric, all entries are strictly positive, and each row and column sums up to one, making $A$ a doubly-stochastic matrix.

This abstraction allows us to leverage the theoretical framework of consensus algorithms (in particular, the convergence properties of row-stochastic averaging processes) to analyze our FL implementation, even though the actual system uses centralized aggregation.

\bigskip
\subsection{FedAvg as One-Shot Consensus}
\label{subsec:fedavg_consensus}

The fully connected and uniformly weighted graph abstraction introduced above, encoded by the row-stochastic matrix $A \in \mathbb{R}^{N\times N}$ with $A_{ij} = 1/N$, underpins the consensus process in our federated setting. While $A$ describes the averaging dynamics among $N$ agents when local values are scalars, each client's model parameter $w_i$ in our context is a vector in $\mathbb{R}^d$. To extend the consensus operation to these vector-valued states, we stack all client parameters into a block vector
\begin{equation}
  x =
  \begin{bmatrix}
    w_1 \\[-2pt] \vdots \\[-2pt] w_N
  \end{bmatrix}
  \in\mathbb{R}^{Nd},
  \qquad
  w_i \in\mathbb{R}^{d}.
  \label{eq:stacked_vec}
\end{equation}

In Federated Averaging, after each local-SGD epoch, the server performs a pure
linear averaging step,
\begin{equation}
  x^{(k+1)}
  =
  \tilde{A} \, x^{(k)},
  \qquad k=0,1,\dots,
  \label{eq:consensus_step}
\end{equation}
where $\tilde{A} \in \mathbb{R}^{N d \times N d}$ is a  doubly-stochastic matrix,  
\begin{equation}
  \tilde{A} = A\otimes I_d,
  \label{eq:A_matrix}
\end{equation}

\noindent where $I_d$ is the $d \times d$ identity and $\otimes$ denotes the Kronecker product, with \(\operatorname{rank}(\tilde A)=\operatorname{rank}(A)\cdot\operatorname{rank}(I_d)=1\cdot d=d\).
The Kronecker product construction $\tilde{A} = A \otimes I_d$ extends the scalar averaging operator $A$ to act naturally on the joint parameter space $\mathbb{R}^{N d}$, where each agent's parameter $w_i \in \mathbb{R}^d$ is stacked into the vector $x$ as in~\eqref{eq:stacked_vec}. Concretely, for any $x$, the action of $\tilde{A}$ can be written as
\[
\tilde{A} x =
\begin{bmatrix}
\sum_{j=1}^N A_{1j} w_j \\
\vdots \\
\sum_{j=1}^N A_{Nj} w_j
\end{bmatrix},
\]
meaning that each agent's new parameter $w_i^{(k+1)}$ is the weighted average of all agents' parameters, exactly as in the scalar case, but now performed for each vector coordinate independently. The Kronecker product with $I_d$ ensures that the same mixing pattern (as encoded by $A$) is applied identically to every component of the parameter vectors, preserving the structure of the averaging operation across all $d$ dimensions. Thus, the averaging step \(x^{(k+1)}=\tilde{A}x^{(k)}\) enforces that, after one application, all \(w_i\) are replaced by their average, coordinate-wise, matching the theoretical result for scalar averaging and guaranteeing synchronous agreement over the entire parameter space. This algebraic construction is standard in consensus and multi-agent systems theory~\cite{colla2022automated, bullo2023lectures}.

These properties follow directly from the definition of $\tilde A$:
(i) $\tilde A$ is doubly-stochastic: every row and column sums to one, so (2) is the classical averaging system $x^{(k+1)} = \tilde A x^{(k)}$ studied in consensus theory.
(ii) $\tilde A$ is an idempotent block-averaging operator of rank $d$: its spectrum is 
$\lambda_1=\cdots=\lambda_d=1$ and $\lambda_{d+1}=\cdots=\lambda_{Nd}=0$,  $\tilde A^{2}=\tilde A$ and consensus is reached in a single step \cite{bullo2023lectures}.

\vspace{1ex}
Because all local models collapse to their arithmetic mean every round, the long-run behavior of FedAvg is determined entirely by the local update preceding~\eqref{eq:consensus_step}. The consensus step itself guarantees instant agreement, while the stochastic gradients steer the trajectory of the shared model across communication rounds.

Equation (2) is a special case of the French--Harary--DeGroot model 
\begin{equation}
  x^{(k+1)}
  =
  \ A \, x^{(k)},
  \qquad k=0,1,\dots,
  \label{eq:consensus_step}
\end{equation} with a doubly-stochastic ${A}$, whose associated digraph is strongly connected. According to Theorem 5.1 in~\cite{bullo2023lectures}, such systems converge to
\(
  \bigl(\tfrac{1}{N}\sum_{i=1}^{N} w_i\bigr)\mathbf 1_N
\)
regardless of initialization, where \(\mathbf{1}_N\) denotes the \(N\)-dimensional vector of all ones. In FedAvg, gradient descent is embedded within this block rank-one averaging process, so that each round consists of a local step,
\begin{align}
  \text{Local step:}\quad  & 
    w_i \leftarrow w_i - \eta \nabla F_i(w_i), \nonumber\\[2pt]
  \text{Consensus:}\quad  &
    x^{(k+1)} = \tilde{A} \, x^{(k)} 
\end{align}

\vspace{1ex}
Mathematically, the composite update can be written as
\begin{equation}
\label{eq:fedavg_composite}
\begin{aligned}[t]          % ← [t] puts the number on the first line
x^{(k+1)} &= \tilde{A}\!\bigl(x^{(k)} - \eta\,g^{(k)}\bigr),\\
g^{(k)}   &= \operatorname{stack}\!\bigl(\nabla F_{1},\ldots,\nabla F_{N}\bigr).
\end{aligned}
\end{equation}

which eliminates disagreement between clients due to heterogeneous gradients while still following the global descent direction, as will be further explored in Section~\ref{num_res}.

\vspace{1ex}
Regarding the guarantee of convergence of FedAvg, under standard assumptions, such as smoothness and bounded variance (as defined in Section II), FedAvg converges at rate $\mathcal{O}(1/ T)$ where T is the number of SGDs~\cite{li2019convergence}.

% Optionally, you can add a remark:
\section{Numerical Examples/Simulations}
\label{num_res}

The forecasting task is the short-term prediction of market volatility trends. In order to deploy the FedAvg algorithm we rely on the open-source federated learning framework Flower \cite{beutel2020flower}. For the analysis we utilize open-source stock market data, specifically focusing on the S\&P 500 index as a baseline for performance evaluation, then we investigate more elaborated scenarios and we use also DAX and FTSE 100 open-source data. Acknowledging the inherently high noise-to-signal ratio characteristic of financial market indices, which typically necessitates extensive, multi-source datasets and large-scale computational models for accurate price prediction, this research adopts an alternative focus. Given limitations in data access, computational resources, training time and as the research primarily focuses on the federated learning aspect, the prediction complexity is heavily reduced to function as a proof of concept. The objective is narrowed to forecasting the phenomenon of volatility clustering. This well-documented dynamic in financial markets describes the tendency for periods of high (or low) volatility to persist.

Feature engineering for the predictive model is performed as follows: for each day within the period of January 1, 2015, to December 31, 2023, the log-return is computed. The log-return for day t is defined as $\log(P_t/P_{t-1})$, where $P_t$ represents the closing price of day $t$. Log-returns are employed due to their established additive and stationary properties in financial modeling. Additionally, cyclical temporal features such as day of the week, month and day of the month are incorporated, with their raw values sine/cosine encoded in order to accurately capture seasonality. To mitigate the impact of high-frequency noise, a 5-day running average is applied to the log-returns, effectively smoothing the data.

The prediction task is formulated as a binary classification problem: classifying each instance as either high volatility (1) or low volatility (0). Specifically, the model is tasked with predicting the volatility for the next day, given a 5-day time horizon of past input data. The threshold distinguishing high from low volatility is determined as the median of the training dataset's volatility distribution. This ensures an approximately balanced dataset, with 50\% of samples classified as low volatility and 50\% as high volatility. Consequently, an accuracy metric exceeding 50\% in the subsequent results indicates the model's ability to discern and predict underlying volatility trends beyond random choice.

All models share the same LSTM architecture, trained with the Adam optimizer and binary cross-entropy loss. The number of federated agents $N$ is scenario-dependent; typically, the historical time series is split 80\% for training and 20\% for testing. Local models are trained only on their assigned training partition, while the centralized baseline is trained on the full training set.

We report both binary cross-entropy (BCE) loss and classification accuracy on the test set at every epoch (centralized/local) or round (federated). Unless otherwise stated, results are visualized as follows: the mean $\pm1\sigma$ band of local-only models (gray), the federated learning trajectory (solid line), and the centralized model (dashed line). Additionally, we report the performance of each training scheme at its best epoch, defined as the epoch with the lowest training loss, showing both the exact BCE and accuracy values. This consistent setup enables direct, transparent comparison across all experimental regimes.

Our primary dataset consists of daily S\&P 500 index data from 2015 to 2023, retrieved from Yahoo Finance. Additional indices (DAX, FTSE~100) are used for transfer and personalization experiments. The code used in our simulations is available at our GitHub repository\footnote{\url{https://github.com/lvb2000/ATIC_FL.git}}.

\subsection{Randomly split data}
\label{first_subsec}
We conduct our first sub-experiment using S\&P 500. Furthermore, the equal-sized partitions assigned to each agent are randomly sampled from the training set, meaning that these partitions do not correspond to any specific time period and that the three partitions are independent, identically distributed (IID). The test dataset is chronologically sorted and never shuffled. 

\begin{figure}[ht]
    \centering
    \includegraphics[width=\linewidth]{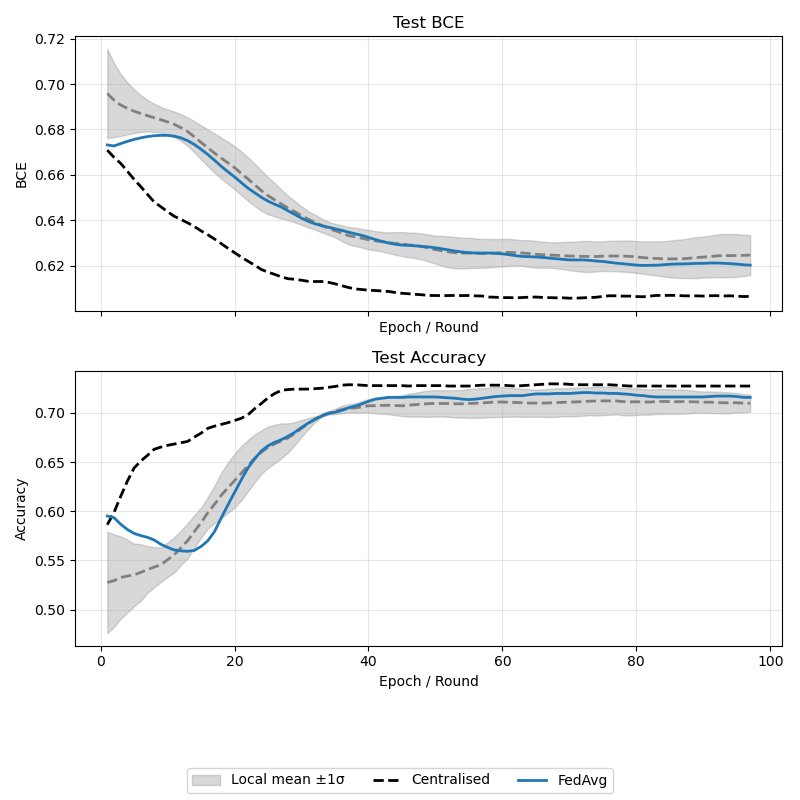}
    \caption{Results for randomly split dataset}
    \label{fig:stacked-figures}
\end{figure}

Fig. \ref{fig:stacked-figures} shows that federated learning consistently under-performs the centralized baseline, but often surpasses (or matches) the average performance of the individual local agents. The federated model has greater ability to generalize, maintaining an improved and more stable test performance throughout training than any individual clients having only a partition of the dataset.
\begin{table}[htbp]
  \centering
  \caption{Performance comparison on randomly split data}
  \label{tab:perf_compare}
  \begin{tabular}{lcc}
    \toprule
    \textbf{Training Scheme}            & \textbf{BCE}                        & \textbf{Accuracy (\%)}               \\
    \midrule
    \textcolor{localgrey}{Local (Mean)}          & \textcolor{localgrey}{0.6261}      & \textcolor{localgrey}{0.7102}       \\
    \textcolor{fedavgblue}{Federated (FedAvg)}   & \textcolor{fedavgblue}{0.6197}     & \textcolor{fedavgblue}{0.7162}      \\
    \textcolor{globaldark}{Centralized (Global)} & \textcolor{globaldark}{0.6052}     & \textcolor{globaldark}{0.7295}      \\
    \bottomrule
  \end{tabular}
\end{table}
Table I shows that FedAvg consistently achieves performance between isolated (Local) and centralized (Global) training. Compared to local-only training, FedAvg reduces the loss (BCE) by approximately 1.0\% and increases accuracy by about 0.8\%. Nevertheless, FedAvg still lags behind the centralized model by c.a. 2.4\% in terms of loss and about 1.3\% in accuracy. Overall, these results confirm that federated collaboration improves performance compared to isolated models while approaching the fully centralized baseline. 

\subsection{Federated Learning with Temporally Partitioned (Non-IID) Data}

In order to study the effect of client data heterogeneity on collaborative forecasting, we repeat the federated learning experiment described previously, but adopt a non-IID data partitioning strategy. Instead of randomly distributing training samples to each agent, we assign data based on calendar quarters: each client receives all training samples corresponding to a specific segment of the year (e.g., January--March, April--June, July--September, or October--December). This means that each client observes only a subset of the possible market regimes and seasonal effects, resulting in statistically heterogeneous (non-IID) local datasets. Such settings are common in real-world federated learning applications, where the data available to each participant may reflect local, temporal, or contextual biases~\cite{kairouz2021advances}.

All model architectures, optimization settings and evaluation protocols are kept identical to the IID baseline. The test set remains chronologically ordered and is shared across all agents for consistent comparison.

\begin{figure}[ht]
    \centering
    \includegraphics[width=\linewidth]{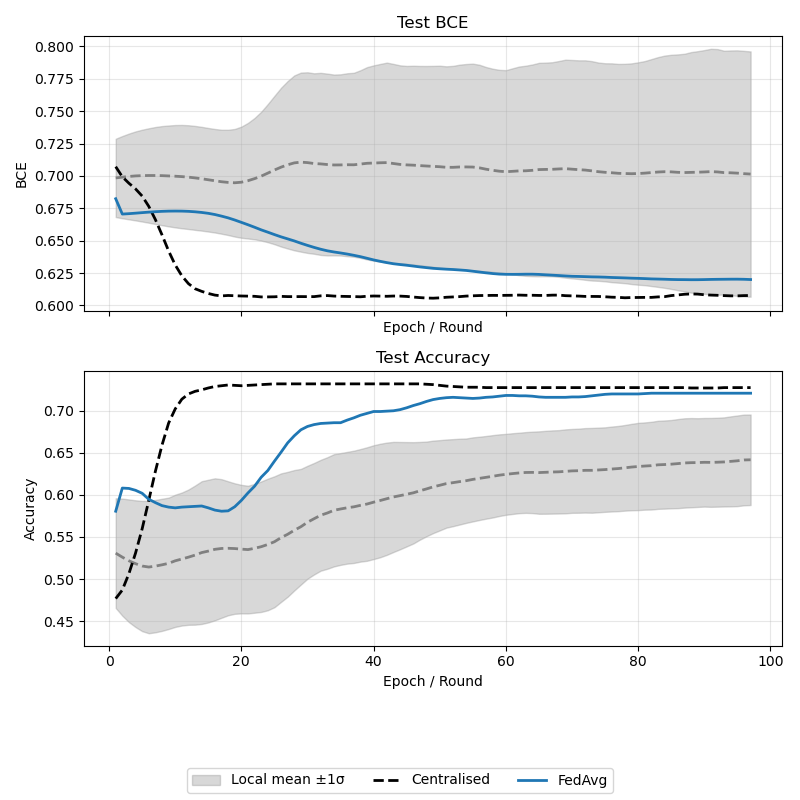}
    \caption{Results for federated learning with temporally partitioned (non-IID) data.}
    \label{fig:quarters-figures}
\end{figure}

Figure~\ref{fig:quarters-figures} illustrates the enhanced performance capabilities of the federated learning approach, particularly when compared to the IID case. It is observed that individual agents, trained solely on seasonal one-fourth partitions of the dataset, exhibit significantly degraded performance relative to a centralized model trained on the complete dataset. Not only is the average performance reduced, but the variance between agents also increases a lot, making the individual predictions less robust. Notably, the federated learning agent's performance closely approximates that of the centralized model, demonstrating substantial improvement over the individual agents. This suggests that the aggregation of weights from individual agents facilitates robust generalization to the test loss landscape minimum, indicating an effective collaborative learning mechanism.

\begin{table}[htbp]
  \centering
  \caption{Performance comparison non-IID}
  \label{tab:perf_compare}
  \begin{tabular}{lcc}
    \toprule
    \textbf{Training Scheme}            & \textbf{BCE}                  & \textbf{Accuracy (\%)}         \\
    \midrule
    \textcolor{localgrey}{Local (Mean)}          & \textcolor{localgrey}{0.7047}    & \textcolor{localgrey}{0.5366}    \\
    \textcolor{fedavgblue}{Federated (FedAvg)}   & \textcolor{fedavgblue}{0.6193}   & \textcolor{fedavgblue}{0.7206}   \\
    \textcolor{globaldark}{Centralized (Global)} & \textcolor{globaldark}{0.6046}   & \textcolor{globaldark}{0.7318}   \\
    \bottomrule
  \end{tabular}
\end{table}

Table II reports that under temporally partitioned (non-IID) data, the performance gap between local and centralized training becomes more pronounced. Local models suffer from both high loss and low accuracy due to limited seasonal exposure. Indeed, FedAvg significantly outperforms local models, reducing the loss by approximately 12\% and improving accuracy by nearly 18.4\%, showing strong benefits from collaborative training. However, it still lags slightly behind the centralized model, which achieves the best performance. Indeed FedAvg’s binary-cross-entropy is 2.4\% higher than the centralized model’s, and its accuracy is 1.5\% lower.  This confirms that while non-IID settings amplify the disadvantages of isolated training, federated learning remains a powerful tool to recover much of the lost performance through aggregation.

\subsection{Federated Learning with Client Heterogeneity}

So far, our discussion has assumed a homogeneous client environment, where each client possesses identical computational resources and performs an equal number of gradient updates per epoch. However, real-world decentralized systems frequently exhibit heterogeneity in client computational capabilities and network connectivity. To investigate the impact of such variations, we designed an experiment where clients selectively skip optimization updates across epochs, thereby creating a scenario with differing training speeds among agents. Specifically, the first agent updates every epoch, the second every two epochs, the third every three epochs, and so on. Aside from this modification, the experimental setup remains consistent with our baseline configuration.

\begin{figure}[ht]
    \centering
    \includegraphics[width=\linewidth]{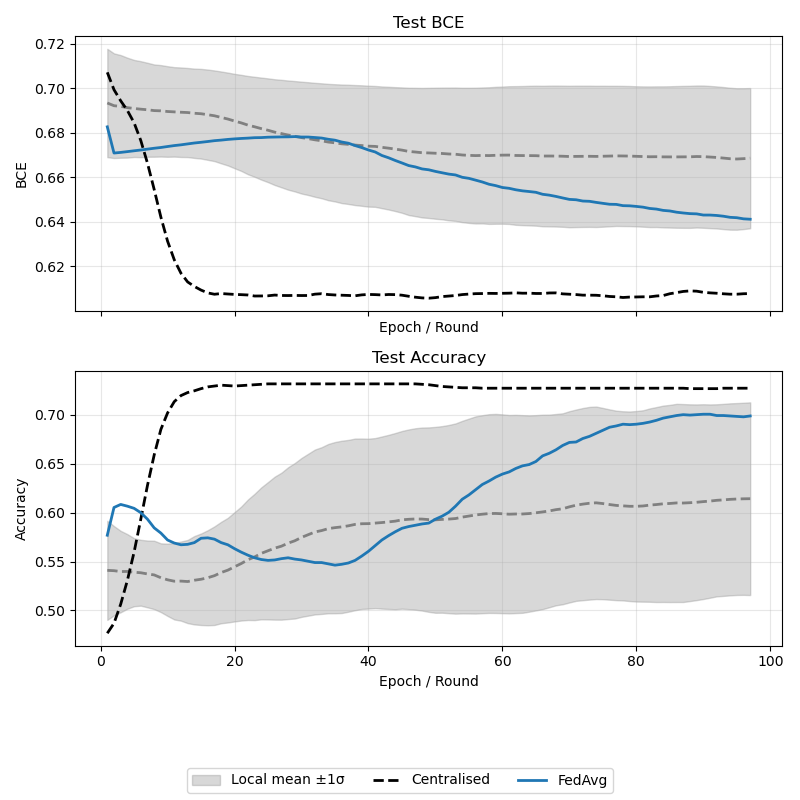}
    \caption{Results for federated learning with client heterogeneity.}
    \label{fig:heterogeneity-figures}
\end{figure}

The results of this experiment are presented in Figure~\ref{fig:heterogeneity-figures}. A notable observation is the drastically increased variance among agents' performance, which directly stems from their heterogeneous training speeds. Compared to the baseline scenario, this heterogeneity also introduces a slowdown in the convergence of the federated average. Nevertheless, despite these significant imbalances, the federated average does not diverge and ultimately achieves a performance level comparable to the homogeneous baseline. This outcome demonstrates the robustness of the federated learning approach even when faced with heavily imbalanced client contributions.

% In your document:
\begin{table}[htbp]
  \centering
  \caption{Performance comparison Epoch Variance}
  \label{tab:perf_compare}
  \begin{tabular}{lcc}
    \toprule
    \textbf{Training Scheme}            & \textbf{BCE}                  & \textbf{Accuracy (\%)}         \\
    \midrule
    \textcolor{localgrey}{Local (Mean)}          & \textcolor{localgrey}{0.6870}    & \textcolor{localgrey}{0.5881}    \\
    \textcolor{fedavgblue}{Federated (FedAvg)}   & \textcolor{fedavgblue}{0.6404}   & \textcolor{fedavgblue}{0.6985}   \\
    \textcolor{globaldark}{Centralized (Global)} & \textcolor{globaldark}{0.6046}   & \textcolor{globaldark}{0.7318}   \\
    \bottomrule
  \end{tabular}
\end{table}
Under the epoch-variance setting, FedAvg remarkably improves over local-only training: it achieves a lower BCE loss by 6.8\% and improves accuracy by  18.8\%. As expected, FedAvg still falls behind the centralized model, with a loss that is  5.9\% higher and an accuracy that is 4.6\% lower than the global benchmark. Overall, aggregation recovers most of the performance lost to asynchronous client updates while remaining slightly below fully pooled training. 

\subsection{Federated Learning with Differential Privacy}

In some cases, even local model update information can be sensitive, so clients may mask it slightly to obtain stronger privacy guarantees. In order to evaluate the impact of strong privacy guarantees on collaborative forecasting, we repeat the randomly partitioned S\&P~500 experiment under the differential privacy (DP) framework. Differential privacy~\cite{dwork2006calibrating} is a mathematical definition of privacy for algorithms analyzing and sharing statistical information about datasets. An algorithm is said to be $(\epsilon, \delta)$-differentially private if the addition or removal of any single data point has a negligible effect on the output distribution, bounded by the privacy parameters $\epsilon$ (privacy loss) and $\delta$ (probability of failure). This is typically accomplished by adding carefully calibrated random noise to intermediate computations or outputs, such as model updates in federated learning. The resulting guarantee ensures that an adversary observing the model cannot confidently infer the presence or absence of any individual in the training data.

In our DP-FL protocol, we inject Gaussian noise into the local model updates communicated by each agent at every round, following standard practice. In this implementation, privacy is controlled by specifying the L2 clipping norm and the Gaussian noise standard deviation added to each client update; the formal $(\epsilon, \delta)$ privacy guarantee is not computed or reported. All other experimental settings (data partition, model architecture, training protocol and test set) are identical to those in Section~\ref{first_subsec}.

It is important to note that, for these results, we deliberately did not tune the privacy or noise parameters to optimize performance. Rather, we chose a high noise level to explicitly demonstrate how excessive noise, intended for strong privacy, can severely degrade model utility: model updates become dominated by noise, overwhelming any useful signal.

\begin{figure}[ht]
    \centering
    \includegraphics[width=\linewidth]{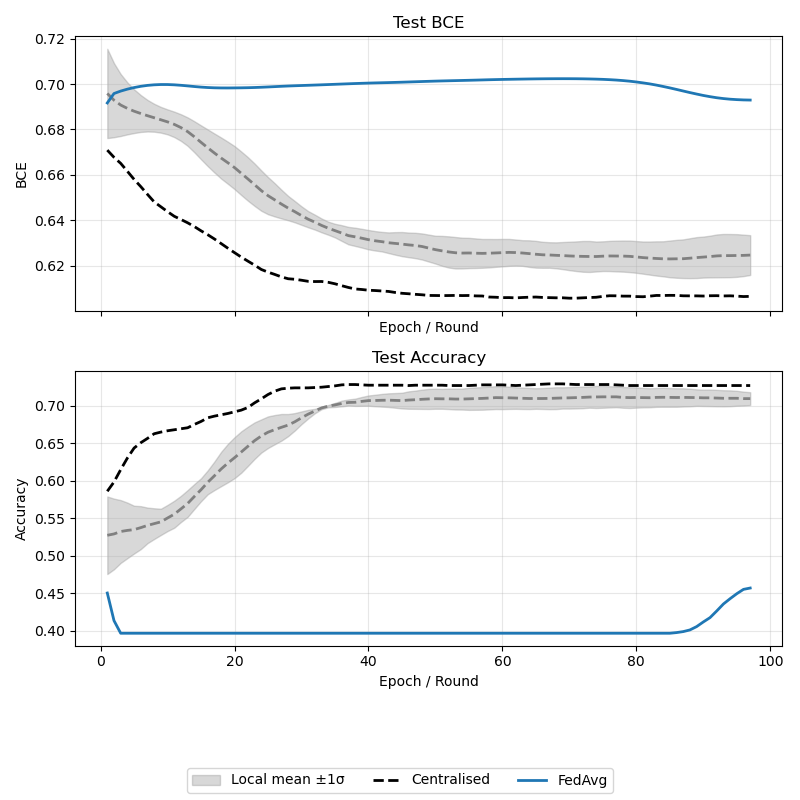}
    \caption{Performance of federated learning with differential privacy. Adding strong DP noise prevents the model from learning, resulting in the poorest performance among all experiments.}
    \label{fig:stacked-figures2}
\end{figure}

As shown in Figure~\ref{fig:stacked-figures2}, the DP-FL model fails to learn meaningful patterns from the data. Both the loss and accuracy curves indicate near-random predictions throughout training. 
\begin{table}[htbp]
  \centering
  \caption{Performance comparison DP}
  \label{tab:perf_compare_dp}
  \begin{tabular}{lcc}
    \toprule
    \textbf{Training Scheme}            & \textbf{BCE}                  & \textbf{Accuracy (\%)}         \\
    \midrule
    \textcolor{localgrey}{Local (Mean)}          & \textcolor{localgrey}{0.6261}    & \textcolor{localgrey}{0.7102}    \\
    \textcolor{fedavgblue}{Federated (FedAvg)}   & \textcolor{fedavgblue}{0.6771}   & \textcolor{fedavgblue}{0.5810}   \\
    \textcolor{globaldark}{Centralized (Global)} & \textcolor{globaldark}{0.6052}   & \textcolor{globaldark}{0.7295}   \\
    \bottomrule
  \end{tabular}
\end{table}

Under the high-noise DP setting, Federated training scheme performs significantly  worse than both baselines. Its loss is  8\% higher than local training  and  12\% higher than the centralized model, while its accuracy drops by about 13\% (18\% relative) with respect to local and by c.a. 15\%  (20\% relative) versus centralized. These evident degradations confirm that if the strong noise added for privacy is not correctly tuned, it might overwhelm useful signal, making federated collaboration not fruitful.
\subsection{Personalized Federated Learning and Dataset Transfer Protocol}

To further investigate model adaptability and the benefits of collaborative learning, we conduct a personalized federated learning (personalized FL) experiment involving two distinct datasets. After federated training on S\&P~500 and DAX data (two agents) and centralized training on their union, we evaluate both (federated and centralized) on a test set that is the union of the S\&P~500 and DAX test splits. This mixed-domain test set allows us to assess the generalization ability of the federated model across both markets.

Following this, we explore model personalization by introducing data from the FTSE~100 index, which was not observed during initial training or testing. We personalize the global federated model to the FTSE~100 data using two standard approaches in personalized FL \cite{kulkarni2023survey}: (i) retraining only the final (output) layer while freezing the rest of the network, and (ii) retraining the entire model from scratch on the FTSE~100 training set. Evaluation on the FTSE~100 test set is performed only during the personalization phase. 

This experimental protocol is designed to address a fundamental question in financial machine learning: \emph{To what extent can a forecasting model trained collaboratively on multiple markets (A and B) generalize to a new, unseen market (C) with minimal additional training?} In practical terms, this scenario reflects the real-world need to deploy models in new environments where data is scarce and rapid adaptation is crucial. By initializing on the federated model trained jointly on S\&P~500 and DAX, and then fine-tuning on FTSE~100 with limited data and computational steps, we can directly measure the efficiency of knowledge transfer across markets. The comparison between fine-tuning only the final layer (low adaptation cost) and retraining the entire model from scratch (high adaptation cost) quantifies the benefit of federated pretraining for rapid deployment in novel markets.

\begin{figure}[ht]
    \centering
    \includegraphics[width=\linewidth]{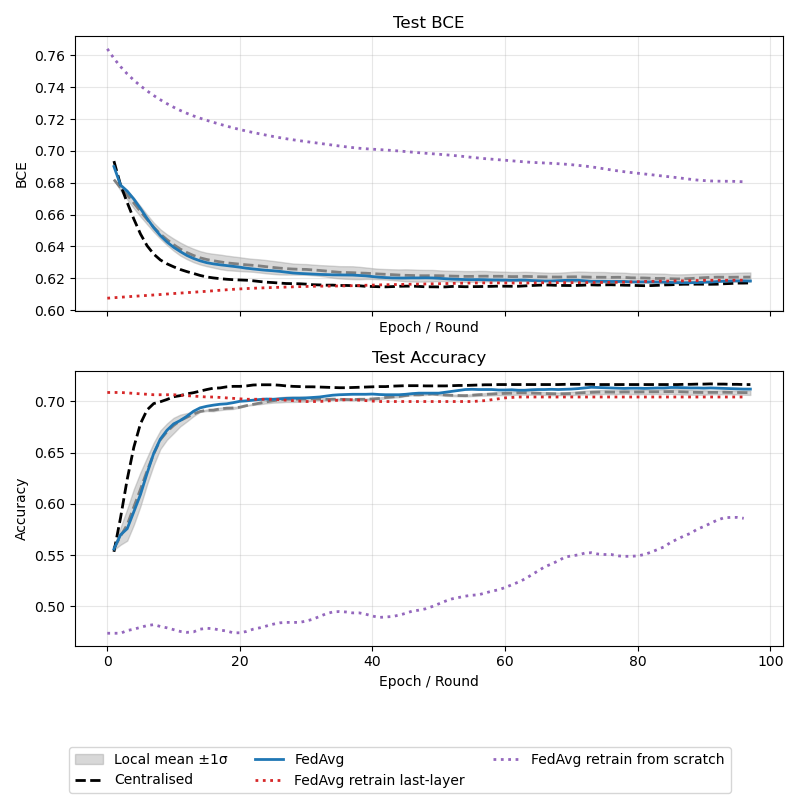}
    \caption{Results for Personalized Federated Learning and Dataset Transfer protocol. Model with FL knowledge outperforms model built from scratch.}
    \label{fig:stacked-figures3}
\end{figure}

The experimental results, shown in Figure~\ref{fig:stacked-figures3}, illustrate the relative performance of these adaptation strategies. 
Two distinct observations emerge from the analysis:

Firstly, when both clients are trained on two separate, complete datasets, their individual performance reaches the level observed for the centralized model trained on the combined dataset. This suggests a potential limitation in the model's capacity or an indication of limited statistical dependencies within the dataset, where simply increasing the volume of training data does not yield further performance improvements. This finding similarly applies to the federated average, which also does not exceed the centralized model's score.

Secondly, evaluating the model's transferability to a novel stock index yielded positive results. Specifically, retraining the model from scratch on a new stock index (FTSE 100), comprising only 10\% of the data size of the original pre-trained dataset, did not enable it to achieve the same performance as fine-tuning only the last layer using the same amount of data and leveraging the weights previously learned by the federated average. This indicates that the learned trends from the initial training generalize effectively to unseen financial domains, and that pre-trained weights from the federated average provide a significant advantage for new tasks, even with reduced data availability.

\begin{table}[htbp]
  \centering
  \caption{Performance comparison Dataset Transfer Protocol}
  \label{tab:perf_compare}
  \begin{tabular}{lcc}
    \toprule
    \textbf{Training Scheme}            & \textbf{BCE}                  & \textbf{Accuracy (\%)}         \\
    \midrule
    \textcolor{localgrey}{Local (Mean)}          & \textcolor{localgrey}{0.6194}    & \textcolor{localgrey}{0.7098}    \\
    \textcolor{fedavgblue}{Federated (FedAvg)}   & \textcolor{fedavgblue}{0.6167}   & \textcolor{fedavgblue}{0.7141}   \\
    \textcolor{globaldark}{Centralized (Global)} & \textcolor{globaldark}{0.6139}   & \textcolor{globaldark}{0.7141}   \\
    \textcolor{red}{Federated + (Last layer)}     & \textcolor{red}{0.6069}          & \textcolor{red}{0.7090}          \\
    \textcolor{localgrey}{Scratch}                & \textcolor{localgrey}{0.6800}    & \textcolor{localgrey}{0.5828}    \\
    \bottomrule
  \end{tabular}
\end{table}
With respect to the local baseline, FedAvg reduces BCE by 0.44\% and increases accuracy by about 0.6 percentage points. Centralized training achieves a slightly better BCE reduction of 0.89\%, but does not improve accuracy beyond FedAvg. Notably, a simple last-layer fine-tuning on the FTSE-100 data lowers BCE by 2.02\% with only a minimal decrease in accuracy. In contrast, training from scratch on the small FTSE-100 split leads to a 9.8\% increase in BCE and a 12.7\% drop in accuracy compared to last-layer fine-tuning. These results highlight the strong transfer-learning advantage provided by starting from a federated pre-trained model.

\iffalse
\medskip

\noindent \textbf{Extensions Toward Realism.} To better reflect practical federated learning scenarios, we extended our experiments with the following features:
\\
- \emph{Data partitioning:} We considered both randomly shuffled (IID) splits and non-IID splits, where each agent was assigned data from distinct, non-overlapping time periods. This allowed us to evaluate the impact of data heterogeneity, as would occur if agents observed different market segments or regimes.
\\
- \emph{Client heterogeneity:} We modeled differences in agent connectivity and computational resources by introducing partial participation (only a random subset of agents participate in each round) and asynchronous updates (agents send updates at different times).
\\
- \emph{Privacy-preserving mechanisms:} Differential privacy was applied to model updates, limiting the risk of sensitive information leakage from individual agents during the federated training process.
\\
- \emph{Personalized federated learning:} Beyond a single global model, we explored partially personalized models, allowing agents to learn both shared and individual components—an approach particularly advantageous under non-IID conditions.

That way, our setup incorporates several key challenges and advancements relevant to realistic federated learning in financial forecasting.
\fi

\section{Results}

The main contribution of this paper is a systematic study of Federated Learning  in the context of binary classification for financial market volatility, under realistic settings like data and clients heterogeneity as well as privacy constraints. We establish the following key findings:

\begin{enumerate}
    \item  Federated learning achieves classification accuracy and generalization performance that match a centralized model trained on the union of all data, while requiring only the exchange of model updates and preserving the privacy of local datasets.
    \item FL significantly outperforms single-agent models trained solely on individual data subsets in terms of both accuracy and robustness, demonstrating the concrete benefit of collaborative learning.
    \item Federated learning demonstrates enhanced generalization on non-IID data distributions, particularly when data is temporally partitioned. In such scenarios, the federated model closely approaches centralized performance and substantially improves upon individual client models, which suffer from increased variance and reduced robustness.
    \item The federated learning approach exhibits remarkable robustness to client heterogeneity in computational resources and training speeds. Despite significant imbalances in client update frequencies, the federated average maintains stable convergence and achieves performance comparable to homogeneous baselines, without divergence.
    \item The model's ability to learn from additional data is constrained by inherent model capabilities and dataset statistical dependencies. While individual clients training on distinct full datasets can match the performance of a centralized model, further increases in data volume do not yield additional performance gains for either individual clients or the federated average.
    \item Knowledge transfer through personalized federated learning significantly boosts performance on novel, unseen datasets. Fine-tuning only the final layer of a federated pre-trained model on a new stock index (FTSE~100) with limited data outperforms retraining the entire model from scratch, highlighting the efficiency of leveraging previously learned weights for rapid adaptation to new market domains.
    \item Strong differential privacy guarantees, when applied without careful tuning, can severely degrade model utility in federated learning. 
\end{enumerate}

\iffalse
    \item These results hold even when (a) additional market features are introduced, (b) agents' data distributions are deliberately made non-IID, (c) personalized FL methods are employed, and (d) differential privacy techniques are applied.  
\fi 

The above findings are supported by extensive numerical experiments on financial time series data using a shared LSTM classifier. The experiments demonstrate that collaborative, privacy-preserving learning provides tangible value in financial applications, even in the presence of practical constraints such as data heterogeneity and the need for personalization.
In this study, we have shown that even in a competitive, collaboration-averse field such as finance, it is possible to achieve improved outcomes without fully disclosing private data. Federated solutions exist and are widely applicable, yielding results that closely match those obtained from centralized, fully pooled data.
We provide a complete pipeline, with code designed to be easily extensible to other financial time series, federated learning research and benchmarking tasks. Federated Learning is a highly promising branch of machine learning. Future research directions could include understanding the optimal privacy–utility trade-off by applying differential privacy techniques, forecasting more complex targets using richer, yet harder-to-access, data sources. Finally, another possibility would be addressing robustness to adversarial or faulty client through robust aggregation techniques or anomaly detection in order to maintain reliability of the framework.

\addtolength{\textheight}{-12cm}   % This command serves to balance the column lengths

%%%%%%%%%%%%%%%%%%%%%%%%%%%%%%%%%%%%%%%%%%%%%%%%%%%%%%%%%%%%%%%%%%%%%%%%%%%%%%%%

%%%%%%%%%%%%%%%%%%%%%%%%%%%%%%%%%%%%%%%%%%%%%%%%%%%%%%%%%%%%%%%%%%%%%%%%%%%%%%%%

%%%%%%%%%%%%%%%%%%%%%%%%%%%%%%%%%%%%%%%%%%%%%%%%%%%%%%%%%%%%%%%%%%%%%%%%%%%%%%%%

%%%%%%%%%%%%%%%%%%%%%%%%%%%%%%%%%%%%%%%%%%%%%%%%%%%%%%%%%%%%%%%%%%%%%%%%%%%%%%%%

{
    \small
    \bibliographystyle{unsrt}
    \bibliography{main}
}

\end{document}